# Learning Bayesian Network Structure from Massive Datasets: The "Sparse Candidate" Algorithm

**Nir Friedman**
Institute of Computer Science
Hebrew University
Jerusalem, 91904, ISRAEL
*nir@cs.huji.ac.il*

**Iftach Nachman**
Center for Neural Computations
Hebrew University
Jerusalem, 91904, ISRAEL
*iftach@cs.huji.ac.il*

**Dana Peér**
Institute of Computer Science
Hebrew University
Jerusalem, 91904, ISRAEL
*danab@cs.huji.ac.il*

## Abstract

Learning Bayesian networks is often cast as an optimization problem, where the computational task is to find a structure that maximizes a statistically motivated score. By and large, existing learning tools address this optimization problem using standard heuristic search techniques. Since the search space is extremely large, such search procedures can spend most of the time examining candidates that are extremely unreasonable. This problem becomes critical when we deal with data sets that are large either in the number of instances, or the number of attributes.

In this paper, we introduce an algorithm that achieves faster learning by restricting the search space. This iterative algorithm restricts the parents of each variable to belong to a small subset of *candidates*. We then search for a network that satisfies these constraints. The learned network is then used for selecting better candidates for the next iteration. We evaluate this algorithm both on synthetic and real-life data. Our results show that it is significantly faster than alternative search procedures without loss of quality in the learned structures.

## 1 Introduction

In recent years there has been a growing interest in learning the structure of Bayesian networks from data [9, 19, 15, 16, 21, 24]. Somewhat generalizing, there are two approaches for finding structure. The first approach poses learning as a *constraint satisfaction* problem. In that approach, we try to estimate properties of conditional independence among the attributes in the data. Usually this is done using a statistical hypothesis test, such as $\chi^2$-test. We then build a network that exhibits the observed dependencies and independencies. Examples of this approach include [21, 24]. The second approach poses learning as an *optimization* problem. We start by defining a statistically motivated *score* that describes the fitness of each possible structure to the observed data. These scores include Bayesian scores [9, 16] and MDL scores [19]. The learner's task is then to find a structure that maximizes the score. In general, this is an NP-hard problem [6], and thus we need to resort to heuristic methods. Although the constraint satisfaction approach is efficient, it is sensitive to failures in independence tests. Thus, the common opinion is that the optimization approach is a better tool for learning structure from data.

Most existing learning tools apply standard heuristic search techniques, such as greedy hill-climbing and simulated annealing to find high-scoring structures. See, for example, [16, 15, 7]. Such "generic" search procedures do not apply any knowledge about the expected structure of the network to be learned. For example, greedy hill-climbing search procedures examine all possible local changes in each step and apply the one that leads to the biggest improvement in score. The usual choice for "local" changes are edge addition, edge deletion, and edge reversal. Thus, there are approximately $O(n^2)$ possible changes where $n$ is the number of variables.[1]

The cost of these evaluations becomes acute when we learn from massive data sets. Since the evaluation of new candidates requires collecting various statistics about the data, it becomes more expensive as the number of instances grows. To collect these statistics, we usually need to perform a pass over the data. Although, recent techniques (e.g., [20]) might reduce the cost of this collection activity, we still expect non trivial computation time for each new set of statistics we need. Moreover, if we consider domains with large number of attributes, then the number of possible candidates grows quickly.

It seems, however, that most of the candidates considered during the search can be eliminated in advance based on our statistical understanding of the domain. For example, in greedy hill-climbing, most possible edge additions might be removed from consideration: If $X$ and $Y$ are almost independent in the data, we might decide not to consider $Y$ as a parent of $X$. Of course, this is a heuristic argument, since $X$ and $Y$ can be marginally independent, yet have strong dependence in the presence of another variable (e.g., $X$ is the XOR of $Y$ and $Z$). In many domains, however, it is reasonable to assume that this pattern of dependencies does not appear.

---

[1] Some of these changes introduce cycles, and thus are not evaluated. Nonetheless, the number of feasible operations is usually quite close to $O(n^2)$.

The idea of using measure of dependence, such as the *mutual information*, between variables to guide network construction is not new. For example, Chow and Liu's algorithm [8] uses the mutual information to construct a tree-like network that maximizes the likelihood score. When we consider networks with larger in-degree, several authors use the mutual information to greedily select parents. However, these authors do not attempt to maximize any statistically motivated score. In fact, it is easy to show situations where these methods can learn erroneous networks. This use of mutual information is a simple example of a statistical cue. In this paper, we incorporate similar considerations within a procedure that explicitly attempts to maximize a score. We provide an algorithm that empirically performs well in massive data sets.

The general idea is quite straightforward. We use statistical cues from the data, to restrict the set of networks we are willing to consider. In this paper, we choose to restrict the possible parents of each variable. Thus, instead of having $n - 1$ potential parents for a variable, we only consider $k$ possible parents, where $k \ll n$. (This is often reasonable, since in many domains we do not expect to learn families with too many parents.) We then attempt to maximize the score with respect to these restrictions. Any search techniques we use in this case will perform faster, since the search space is significantly restricted. Moreover, as we show, in some cases we can find the best scoring network satisfying these constraints. In other cases, we can use the constraints to improve our heuristics.

Of course, such a procedure might fail to find a high-scoring network: a misguided choice of candidate parents in the first phase can lead to a low scoring network in the second phase, even if we manage to maximize the score with respect to these constraints. The key idea of our algorithm is that we use the network we found at the end of the second stage to find better candidate parents. We then can find a better network with respect to these new restrictions. We iterate in this manner until convergence.

The rest of the paper is organized as follows. In Section 2, we review the necessary background on learning Bayesian network structure. In Section 3 we outline the structure of our "sparse candidate" algorithm and show that there are two orthogonal issues that need to be resolved: how to select candidates in each iteration, and how to search given the constraints on the possible parents. We examine these issues in Sections 4 and 5, respectively. In Section 6 we evaluate the performance of the algorithm on synthetic and real-life datasets. We conclude with a discussion of related work and future directions in Section 7

## 2   Background: Learning Structure

Consider a finite set $\mathcal{X} = \{X_1, \ldots, X_n\}$ of discrete random variables where each variable $X_i$ may take on values from a finite set, denoted by $\text{Val}(X_i)$. We use capital letters, such as $X, Y, Z$, for variable names and lowercase letters $x, y, z$ to denote specific values taken by those variables. Sets of variables are denoted by boldface capital letters $\mathbf{X}, \mathbf{Y}, \mathbf{Z}$, and assignments of values to the variables in these sets are denoted by boldface lowercase letters $\mathbf{x}, \mathbf{y}, \mathbf{z}$.

A *Bayesian network* is an annotated directed acyclic graph that encodes a joint probability distribution over $\mathcal{X}$. Formally, a Bayesian network for $\mathcal{X}$ is a pair $B = \langle G, \Theta \rangle$. The first component, namely $G$, is a directed acyclic graph whose vertices correspond to the random variables $X_1, \ldots, X_n$. The graph encodes the following set of conditional independence assumptions: each variable $X_i$ is independent of its non-descendants given its parents in $G$. The second component of the pair, $\Theta$, represents the set of parameters that quantifies the network. It contains a parameter $\theta_{x_i|\mathbf{pa}(X_i)} = P(x_i|\mathbf{pa}(X_i))$ for each possible value $x_i$ of $X_i$, and $\mathbf{pa}(X_i)$ of $\mathbf{Pa}(X_i)$. Here $\mathbf{Pa}(X_i)$ denotes the set of parents of $X_i$ in $G$ and $\mathbf{pa}(X_i)$ is a particular instantiation of the parents. If more than one graph is discussed then we use $\mathbf{Pa}^G(X_i)$ to specify $X_i$'s parents in graph G. A Bayesian network $B$ specifies a unique joint probability distribution over $\mathcal{X}$ given by: $P_B(X_1, \ldots, X_n) = \prod_{i=1}^{n} P_B(X_i|\mathbf{Pa}(X_i))$.

The problem of learning a Bayesian network can be stated as follows. Given a *training set* $D = \{\mathbf{x}^1, \ldots, \mathbf{x}^N\}$ of instances of $\mathcal{X}$, find a network $B$ that *best matches* $D$. The common approach to this problem is to introduce a scoring function that evaluates each network with respect to the training data, and then to search for the best network according to this score. The two scoring functions most commonly used to learn Bayesian networks are the *Bayesian scoring* metric, and the one based on the principle of *minimal description length* (MDL). For a full description see [9, 16] and [3, 19].

An important characteristic of the MDL score and the Bayesian score (when used with a certain class of *factorized* priors, such as the *BDe priors* [16]), is their *decomposability* in presence of full data. When all instances $\mathbf{x}^\ell$ in $D$ are *complete*—that is, they assign values to all the variables in $\mathcal{X}$—the above scoring functions can be decomposed in the following way:

$$\text{Score}(G : D) = \sum_i \text{Score}(X_i \mid \mathbf{Pa}(X_i) : N_{X_i, \mathbf{Pa}(X_i)})$$

where $N_{X_i, \mathbf{Pa}(X_i)}$ are the *statistics* of the variables $X_i$ and $\mathbf{Pa}(X_i)$ in $D$—i.e., the number of instances in $D$ that match each possible instantiation $x_i$ and $\mathbf{pa}(X_i)$.

This decomposition of the scores is crucial for learning structure. A *local* search procedure that changes one arc at each move can efficiently evaluate the gains made by this change. Such a procedure can also reuse computations made in previous stages to evaluate changes to the parents of all variables that have not been changed in the last move. An example of such a procedure is a greedy hill-climbing procedure that at each step performs the local change that results in the maximal gain, until it reaches a local maximum. Although this procedure does not necessarily find a global maximum, it does perform well in practice; e.g., see [16]. Example of other search procedures that advance in one-arc changes include beam-search, stochastic hill-climbing, and simulated annealing.

**Input:**

- A data set $D = \{\mathbf{x}^1, \ldots, \mathbf{x}^N\}$,
- An initial network $B_0$,
- A decomposable score
  $\text{Score}(B \mid D) = \sum_i \text{Score}(X_i \mid \mathbf{Pa}^B(X_i), D)$,
- A parameter $k$.

**Output:** A network $B$.

Loop for $n = 1, 2, \ldots$ until convergence

   **Restrict**

   Based on $D$ and $B_{n-1}$, select for each variable $X_i$ a set $C_i^n$ ($|C_i^n| \leq k$) of candidate parents.
   This defines a directed graph $H_n = (\mathcal{X}, E)$, where $E = \{X_j \rightarrow X_i | \forall i, j, X_j \in C_i^n\}$.
   (Note that $H_n$ is usually cyclic.)

   **Maximize**

   Find network $B_n = \langle G_n, \Theta_n \rangle$ maximizing $\text{Score}(B_n \mid D)$ among networks that satisfy $G_n \subset H_n$ (i.e., $\forall X_i, \mathbf{Pa}^{G_n}(X_i) \subseteq C_{i,}^n$).

Return $B_n$

Figure 1: Outline of the *Sparse Candidate* algorithm

Any implementation of these search methods involves caching of computed counts to avoid unnecessary passes over the data. This cache also allows us to *marginalize* counts. Thus, if $N_{X,Y}$ is in the cache, we can compute $N_X$ by summing over values of $Y$. This is usually much faster than making a new pass over the data. One of the dominating factors in the computational cost of learning from complete data is the number of passes actually made over the training data. This is particularly true when learning from very large training sets.

## 3   The "Sparse Candidate" Algorithm

In this section we outline the framework for our *Sparse Candidate* algorithm The underlying principle for our algorithm is fairly intuitive. It calls for two variables with a "strong dependency" between them to be located "near" each other in the network. The strength of dependency between variables can often be measured using mutual information or correlation [11]. In fact, when restricting the network graph to a tree, Chow and Liu's algorithm [8] does exactly that. It measures the mutual information (formally defined below) between all pairs of variables and selects a maximal spanning tree as the required network.

We aim to use a similar argument for finding networks that are not necessarily trees. Here, the general problem is NP-hard [5]. However, a seemingly reasonable heuristic is to select pairs $(X, Y)$ with high dependency between them and create a network with these edges.

This approach however, does not take more complex interactions into account. For example, if the "true" structure includes a substructure of the form $X \rightarrow Y \rightarrow Z$, we might expect to observe a strong dependency between $X$ and $Y$, $Y$ and $Z$, and also between $X$ and $Z$. However, once we consider both $X$ and $Y$ as parents of $Z$, we might recognize that $X$ does not help in predicting $Z$ once we take $Y$ into account.

Our approach is based on the same basic intuition of using mutual information, but we do so in a refined manner. We use measures of dependency between pairs of variables to *focus* our attention during the search. For each variable $X$, we find a set of variables $Y_1, \ldots, Y_k$ that are the most promising *candidate* parents for $X$. We then *restrict* our search to networks in which only these variables can be parents of $X$. This gives us a smaller search space in which we can hope to find a good structure quickly.

The main drawback of this procedure is, that once we choose the candidate parents for each variable, we are committed to them. Thus, a mistake in this initial stage can lead us to find an inferior scoring network. We therefore iterate the basic procedure, using the constructed network to reconsider the candidate parents and choose better candidates for the next iteration. In the example of $X \rightarrow Y \rightarrow Z$, $X$ would not be chosen as a candidate for $Z$, allowing a variable with weaker dependency to replace it.

The resulting procedure has the general form shown in Figure 1. This framework defines a whole class of algorithms, depending on how we choose the candidates in the **Restrict** step, and how we perform the search in the **Maximize** step. The choice of methods for these two steps are mostly independent of one another. We examine each of these in detail in the next two sections.

Before we go on to discuss these issues, we address the convergence properties of these iterations. Clearly, at this abstract level, we cannot say much about the performance of the algorithm. However, we can easily ensure its monotonic improvement. We require that in the **Restrict** step, the selected candidates for $X_i$'s parents include $X_i$'s current parents, i.e., the selection must satisfy $\mathbf{Pa}^{G_n}(X_i) \subseteq C_i^{n+1}$ for all $X_i$.

This requirement implies that the winning network $B_n$ is a legal structure in the $n + 1$ iteration. Thus, if the search procedure at the **Maximize** step also examines this structure, it must return a structure that scores at least as well as $B_n$. Immediately, we get that $\text{Score}(B_{n+1} \mid D) \geq \text{Score}(B_n \mid D)$.

Another issue is the stopping criteria for our algorithm. There are two types of stopping criteria: a *score based* criterion that terminates when $\text{Score}(B_n) = \text{Score}(B_{n-1})$, and a *candidate based* criterion that terminates when $C_i^n = C_i^{n-1}$ for all $i$. Since the score is a monotonically increasing bounded function, the score based criterion is guaranteed to stop. However, the candidate based criterion might be able to continue to improve after an iteration with no improvement in the score. It can also enter a non-terminating cycle, therefore we need to limit the number of iterations with no improvement in the score.

## 4   Choosing Candidate Sets

In this section we discuss possible measures for choosing the candidate set. To choose candidate parents for $X_i$, we assign each $X_j$ some measure of relevance to $X_i$. As the candidate set of $X_i$, we choose those variables with the highest measure. This general outline is shown in Figure 2. It is clear that in some cases, such as XOR relations, pairwise scoring functions are not enough to capture the dependency between variables. However, for computational efficiency we limit ourselves to this type of functions.

When considering each candidate, we essentially assume that there are no spurious independencies in the data. More precisely, if $Y$ is a parent of $X$, then $X$ is not independent (or "almost" independent) of $Y$, given only a subset of the other parents.

A simple and natural measure of dependence is *mutual information*:

$$I(X;Y) = \sum_{x,y} \hat{P}(x,y) \log \frac{\hat{P}(x,y)}{\hat{P}(x)\hat{P}(y)}$$

Where $\hat{P}$ denotes the observed frequencies in the dataset. The mutual information is always non-negative. It is equal to 0 when $X$ and $Y$ are independent. The higher the mutual information, the stronger the dependence between $X$ and $Y$.

Researchers have tried to construct networks based on $I(X;Y)$, i.e., add edges between variables with high mutual information [8, 12, 22]. While in many cases mutual information is a good first approximation of the candidate parents, there are simple cases for which this measure fails.

**Example 4.1 :** Consider a network with 4 variables $A, B, C,$ and $D$ such that $B \to A$, $C \to A$, $D \to C$. We can easily select parameters for this network such that $I(A;C) > I(A;D) > I(A;B)$. Thus, if we select only two parents based on mutual information, we would select $C$ and $D$. These two, however, are redundant since once we know $C$, $D$ adds no new information about $A$. Moreover, this choice does not take into account the effect of $B$ on $A$. ∎

This example shows a general problem in pairwise selection, which our iterative algorithm overcomes. After we select $C$ and $D$ as candidates, and the learning procedure hopefully only sets $C$ as a parent of $A$, we reestimate the relevance of $B$ and $D$ to $A$. How can this be done with the mutual information? We outline two possible approaches:

The first approach is based on an alternative definition of the mutual information. We can define the mutual information between $X$ and $Y$ as the distance between the distribution $\hat{P}(X,Y)$ and the distribution $\hat{P}(X)\hat{P}(Y)$, which assumes $X$ and $Y$ are independent:

$$I(X;Y) = D_{KL}(\hat{P}(X,Y)\|\hat{P}(X)\hat{P}(Y))$$

where $D_{KL}(P\|Q)$ is the *Kullback-Leibler divergence*, defined as:

$$D_{KL}(P(X)\|Q(X)) = \sum_X P(X) \log \frac{P(X)}{Q(X)}.$$

Thus, the mutual information measures the error we introduce if we assume that $X$ and $Y$ are independent. If we

**Input:**
- Data set $D = \{\mathbf{x}^1, \ldots, \mathbf{x}^N\}$,
- A network $B_n$,
- a score
- parameter $k$.

**Output:** For each variable $X_i$ a set of candidate parents $C_i$ of size $k$.

Loop for each $X_i$ $i = 1, \ldots, n$

- Calculate $M(X_i, X_j)$ for all $X_j \neq X_i$ such that $X_j \notin \mathbf{Pa}(X_i)$
- Choose $x_1, \ldots, x_{k-l}$ with highest ranking, where $l = |\mathbf{Pa}(X_i)|$.
- Set $C_i = \mathbf{Pa}(X_i) \cup \{x_1, \ldots, x_{k-l}\}$

Return $\{C_i\}$

Figure 2: Outline of the *Restrict* step

already have an estimate of a network $B$, we can use a similar test to measure the *discrepancy* between our estimate $P_B(X,Y)$ and the empirical estimate $\hat{P}(X,Y)$. We define

$$M_{\text{Disc}}(X_i, X_j \mid B) = D_{KL}(\hat{P}(X_i, X_j)\|P_B(X_i, X_j))$$

Notice that when $B_0$ is an empty network, with parameters estimated from the data, we get that $M_{\text{Disc}}(X, Y \mid B_0) = I(X:Y)$. Thus, our initial iteration in this case uses mutual information to select candidates. Later iterations use the discrepancy to find variables for which our modeling of their joint empirical distribution is poor. In our example, we would expect that $P_B(A,B)$ in the network, when only $C$ is a parent of $A$, is quite different from $\hat{P}(A,B)$. Thus, $B$ would measure highly relevant to $A$, while $P_B(A,D)$ would be a good approximation of $\hat{P}(A,D)$. Therefore, even "weak" parents have the opportunity to become candidates at some point.

One of the issues with this measure is that it requires us to compute $P_B(X_i, X_j)$ for pairs of variables. When learning networks over large number of variables this can be computationally expensive. However, we can easily approximate these probabilities by using a simple sampling approach. Unlike computation of posterior probabilities given evidence, the approximation of such prior probabilities is not hard. We simply sample $N$ instances from the network, and from these we can estimate all pair-wise interactions. (In our experiments we use $N = 1000$.)

The second approach to extend the mutual information score is based on the semantics of Bayesian networks. Recall that in a Bayesian network $X_i$'s parents *shield* it from its non-descendants. This suggests that we measure whether the conditional independence statement "$X_i$ is independent of $X_j$ given $\mathbf{Pa}(X_i)$" holds. If it holds, then the current parents separate $X_j$ from $X_i$ and $X_j$ is not a parent of $X_i$. On the other hand, if it does not hold, then either $X_j$ is a parent of $X_i$, or $X_j$ is a descendant of $X_i$.

Instead of testing whether the conditional independence statement holds or not, we estimate how strongly it is vio-lated. The natural extension of mutual information for this task, is the notion of *conditional mutual information*:

$$I(X;Y|Z) = \sum_Z \hat{P}(Z) D_{KL}(\hat{P}(X,Y|Z) \| \hat{P}(X|Z)\hat{P}(Y|Z)).$$

This measures the error we introduce by assuming that $X$ and $Y$ are independent given different values of $Z$. We define

$$\mathbf{M}_{\text{Shield}}(X_i, X_j \mid B) = I(X_i; X_j | \mathbf{Pa}(X_i))$$

Once again, we have that if $B_0$ is the empty network, then this measure is equivalent to $I(X_i; X_j)$. Although shielding can remove $X$'s ancestors from the candidate set, it does not "shield" $X$ from its descendants.

A deficiency of both these measures is that they do not take into account the cardinality of various variables. For example if both $Y$ and $Z$ are possible candidate parents of $X$, but $Y$ has two values (one bit of information), while $Z$ has eight values (three bits of information), we would expect that $Y$ is less informative about $X$ than $Z$. On the other hand, we can estimate $P(X|Y)$ more robustly than $P(X|Z)$ since it involves fewer parameters.

Such considerations lead us to use scores which penalize structures with more parameters, when searching the structure space, since the more complex the model is, the easier we are misled by the empirical distribution. We use the same considerations to design such a score for the Restrict step.

To see how to define a measure of this form, we start by reexamining the shielding property. Using the chain rule of mutual information:

$$I(X_i; X_j | \mathbf{Pa}(X_i)) = I(X_i; X_j, \mathbf{Pa}(X_i)) - I(X_i; \mathbf{Pa}(X_i))$$

That is, the conditional mutual information is the additional information we get by predicting $X_i$ using $X_j$ and $\mathbf{Pa}(X_i)$, compared to our prediction using $\mathbf{Pa}(X_i)$. Since the term $I(X_i; \mathbf{Pa}(X_i))$ does not depend on $X_j$, we don't need to compute it when we compare the information that different $X_j$'s provide about $X_i$. Thus, an equivalent comparative measure is

$$\mathbf{M}_{\text{Shield}}(X_i, X_j \mid B) = I(X_i; X_j, \mathbf{Pa}(X_i))$$

Now, if we consider the score of the Maximize step as cautious approximation of the mutual information, with a penalty on the number of parameters, we can get the *score* measure:

$$\mathbf{M}_{\text{Score}}(X_i, X_j \mid B) = \text{Score}(X_i; X_j, \mathbf{Pa}(X_i), D).$$

This simply measures the score when adding $X_j$ to the current parents of $X_i$.

Calculating $\mathbf{M}_{\text{Shield}}$ and $\mathbf{M}_{\text{Score}}$ is more expensive than calculating $\mathbf{M}_{\text{Disc}}$. $\mathbf{M}_{\text{Disc}}$ only needs the joint statistics for all pairs $X_i$ and $X_j$. These require only one pass over the data and the computation can be cached for later iterations. The other measures require the joint statistics of $X_i$, $X_j$, and $\mathbf{Pa}(X_i)$. In general $\mathbf{Pa}(X_i)$ changes between iterations, and usually requires a new pass over the data set each iteration. The cost of calculating these new statistics can be reduced by limiting our attention to variables $X_j$ that have large enough mutual information with $X_i$. Note that this mutual information can be computed using previously collected statistics

## 5 Learning with Small Candidate Sets

In this section we examine the problem of finding a constrained Bayesian network attaining a maximal score. We first show why the introduction of candidate sets improves the efficiency of standard heuristic techniques, such as greedy hill-climbing. We then suggest an alternative heuristic "divide and conquer" paradigm that exploits the sparse structure of the constrained graph.

Formally, we attempt to solve the following problem:

**Maximal Restricted Bayesian Network (MRBN)**
**Input:**

- A set $D = \{\mathbf{x}^1, \ldots, \mathbf{x}^N\}$ of instances
- A digraph $H$ of bounded in-degree $k$
- A decomposable score $S$

**Output:** A network $B = \langle G, \Theta \rangle$ so that $G \subseteq H$, that maximizes S with respect to $D$.

As can be expected, this problem has a hard combinatorial aspect.

**Proposition 5.1:** *MRBN is NP-hard*.

This follows from a slight modification of the NP-hardness of finding an optimal unconstrained Bayesian network [6].

### 5.1 Standard Heuristics

Though MRBN is NP-hard, even standard heuristics are computationally more efficient and give a better approximation compared to the unconstrained problem. This is due to the fact that the search space is substantially smaller, as is the complexity of each iteration, and the number of counts needed.

The search space of possible Bayesian networks is extremely large. By searching in a smaller space, we can hope to have a better chance of finding a high-scoring network. Although the search space size for MRBN remains exponential, it is tiny in comparison to the space of all Bayesian networks on the same domain. To see this, note that even if we restrict the search to Bayesian networks with at most $k$ parents, there are $O(\binom{n}{k})$ possible parent sets for each variable. On the other hand, in MRBN, we have only $O(2^k)$ possible parent sets for each variable. (Of course, the acyclicity constraints disallow many of these networks, but it does not change the order of magnitude in the size of the sets).

Examining the time complexity for each iteration in heuristic searches also points in favor of MRBN. In greedy hill climbing the score for the $O(n^2)$ initial changes are calculated, after which each iteration requires $O(n)$ new calculations. In MRBN we begin with $O(kn)$ initial calculations after which each iteration only requires $O(k)$ calculation.

A large fraction of the learning time involves collecting the sufficient statistics from the data. Here again, restricting to candidate sets saves time. When $k$ is reasonably small, we can compute the statistics for $\{X_i\} \cup C_i$ in one pass over the input. All the statistics we need for evaluating subsets of $C_i$ as parents of $X_i$ can then be computed by marginalization from these counts. Thus, we can dramatically reduce the number of statistics collected from the data.

### 5.2 Divide and Conquer Heuristics

In this section we describe algorithms that utilize the combinatorial properties of the candidate graph $H$ in order to efficiently find the maximal scoring network, given the constraints. To simplify the following discussion, we abstract the details of the Bayesian network learning problem and focus on the underlying combinatorial problem. This problem is specified as follows:

**Input:** A digraph $H = \{X_j \to X_i : X_j \in C_i\}$, and a set of weights $w(X_i, Y)$ for each $X_i$ and $Y \subseteq C_i$.

**Output:** An acyclic subgraph $G \subseteq H$ that maximizes

$$W_H[G] = \sum_i w(X_i, \mathbf{Pa}^G(X_i)).$$

One of the most effective paradigms for designing algorithms is "Divide and Conquer". In this particular problem, the global constraint we need to satisfy is acyclicity. Otherwise, we would have selected, for each variable $X_i$, the parents that attain maximal weight. Thus, we want to decompose the problem into components, so that we can efficiently combine their maximal solutions. We use standard graph decomposition methods to decompose $H$. Once we have such a decomposition, we can find acyclic solutions in each component and combine them into a global solution.

### 5.3 Strongly Connected Components: (SCC)

The simplest decomposition of this form is one that disallows cycles between components, i.e, *strongly connected components*. A subset of vertices $\boldsymbol{A}$ is *strongly connected* if for each $X, Y \in \boldsymbol{A}$, $H$ contains a directed path from $X$ to $Y$ and a directed path from $Y$ to $X$. The set $\boldsymbol{A}$ is *maximal* if there is no strongly connected superset of $\boldsymbol{A}$. It is clear that two maximal strongly connected components must be disjoint, and there cannot be a cycle that involves vertices in both of them (for otherwise their union would be a strongly connected component). Thus, we can partition the vertices in $H$ into maximal strongly connected components. Every cycle in $H$ will be contained within a single component. Thus, once we ensure acyclicity "locally" within each component, we get an acyclic solution over all the variables. This means we can search for a maximum on each component independently.

To formalize this idea, we begin with some definitions. Let $\boldsymbol{A}_1, \ldots \boldsymbol{A}_m$ be a partition of $\{X_1, \ldots, X_n\}$. We define the following subgraphs: $H_{X_i} = \{Y \to X_i | Y \in C_i\}$, $H_j = \bigcup_{X_i \in \boldsymbol{A}_j} H_{X_i}$. For $G \subset H_j$, let $W_{\boldsymbol{A}_j}[G] = \sum_{X_i \in \boldsymbol{A}_j} w(X_i, \mathbf{Pa}^G(X_i))$.

**Proposition 5.2 :** *For $\boldsymbol{A}_1, \ldots, \boldsymbol{A}_m$ strongly connected components of $H$, if for each $j$, $G_j \subset H_j$ is the acyclic graph that maximizes $W_{\boldsymbol{A}_j}[G]$ then*

- *The graph $G = \cup_j G_j$ is acyclic.*
- *$G$ maximizes $W_H[G]$.*

Decomposing $H$ into strongly connected components takes linear time (e.g., see [10]), therefore we can apply this decomposition, and search for the maxima on each component separately. However, when the graph contains large connected components, we still face a hard combinatorial problem of finding the graphs $G_j$. For the remainder of this section we will focus on further decomposition of such components.

### 5.4 Separator Decomposition

We now decompose strongly connected graphs, therefore we must consider cycles between the components. However, our goal is to find small "bottlenecks" through which these cycles must go through. We then consider all possible ways of breaking the cycles at these bottlenecks.

**Definition 5.3:** A *separator* of $H$ is a set $S$ of vertices so that:

1. $H \setminus S$ has two components $H'_1$ and $H'_2$ with no edges between them. For $j \in \{1, 2\}$ let $H_i = H'_i \cup S$.
2. For each $X_i, \exists j \in \{1, 2\}$ so that $\{X_i \cup C_i\} \subseteq H_j$

∎

For each vertex we search for the maximal choice of parents in only one component ($H_1$ or $H_2$). Let $A_1$ and $A_2$ be a disjoint partition of all vertices into two sets, so that if $X_i \in A_j$, then $X_i \cup C_i \subset H_j$. The second property of the separator ensures that such a partition exists. This property holds when $S$ "separates" the moralized graph of $H$, (where each $X_i \cup C_i$ appear as a clique) into two components.

Unlike the SCC decomposition, however, this decomposition does not allow us to maximize $W$ for each $H_j$ independently. Suppose that we find two acyclic graphs $G_1$ and $G_2$ that maximize $W_{A_1}[]$ and $W_{A_2}[]$, respectively. If the combined graph $G = G_1 \cup G_2$ is acyclic, then it must maximize $W_H[]$. Unfortunately, $G$ might be cyclic. The first property of separators ensures that the source of potential conflicts between $G_1$ and $G_2$ involve vertices in the separator $S$.

For $X, Y \in S$, if there is a path from $X$ to $Y$ in $G_1$ and in addition there is a path from $Y$ to $X$ in $G_2$, then the combined graph will be cyclic. Conversely, it is also easy to verify, that any cycle in $G$ must involve at least two vertices in $S$.

This suggests a way of ensuring that the combined graph will be acyclic. If we force some order on the vertices in $S$, and require both $G_1$ and $G_2$ to respect this order, then

**Separator-Algorithm**

- for of each possible order $\sigma$ on $S$
  - For each $i = 1, 2$, find $G_{i,\sigma} \subset H_i$, that maximizes $W_{H_i}[G]$ among graphs that respect $\sigma$.
  - let $G_\sigma = G_{1,\sigma} \cup G_{2,\sigma}$
- Return $G = \arg\max_{G_\sigma} W[G_\sigma]$.

Figure 3: Outline of using a separator to efficiently solve MRBN

we disallow cycles. Formally, let $\sigma$ be a partial order on $\{X_1, \ldots, X_n\}$. We say that a graph $G$ *respects* $\sigma$, if whenever there is a directed path $X_j \to \ldots \to X_i$ in $G$, then $X_i \not\prec_\sigma X_j$.

**Proposition 5.4:** *Let $S$ be a separator in $H$ and let $\sigma$ be a complete order on $S$. Let $G_1 \subset H_1$ and $G_2 \subset H_2$ be two acyclic graphs that respect $\sigma$. Then, $G = G_1 \cup G_2$ is acyclic.*

Given $S$, a small separator in $H$, this suggests a simple algorithm described in figure 3. This approach considers $|S|!$ pairs of independent sub-problems. If the cost of finding a solution to each of the sub-problems is smaller than for the whole problem, and $|S|$ is relatively small, this procedure can be more efficient.

**Proposition 5.5:** *Using the same notation as in the separator-algorithm, if $\forall \sigma$ for $j \in \{0, 1\}$, $G_{j,\sigma}$ maximizes $W_{H_j}[]$ among the graphs that respect $\sigma$ then:*

- *$G_\sigma$ maximizes $W_H[]$ among the graphs that respect $\sigma$*
- *$G = \arg\max_{G_\sigma} W[G_\sigma]$ maximizes $W_H[]$.*

Proposition 5.5 implies that algorithm 3 returns the optimal solution.

### 5.5 Cluster-Tree Decomposition

In this section we present *cluster trees*, which are representations of the candidate graphs, implying a recursive separator decomposition of $H$ into *clusters*. The idea is similar to those of standard clique-tree algorithms used for Bayesian network inference (e.g., [17]). We use this representation to discuss a class of graphs for which $W_H[]$ can be found in polynomial time.

**Definition 5.6:** A *Cluster Tree* of $H$ is a pair $(U, T)$, where $T = (J, F)$ is a tree and $U = \{U_j | j \in J\}$ is a family of *clusters*, subsets of $\{X_1, \ldots, X_n\}$, one for each vertex of $T$, so that:

- For each $X_i$, there exists $j(i) \in J$ such that $\{X_i \cup C_i\} \subseteq U_{j(i)}$.
- For all $i, j, k \in J$, if $j$ is on the path from $i$ to $k$ in $T$, then $U_i \cap U_k \subset U_j$. This is called the *running intersection property*.

∎

We introduce some notation: Let $(i, j)$ be an edge in $T$. Then $S_{i,j} = U_i \cap U_j$ is a separator in $T$, breaking it into two subtrees $T_1$ and $T_2$. Define $A_j$ to be the set of vertices assigned (with their parents) to $U_j$: $A_j = \{X_i | j(i) = j\}$. Define $A[T_i] = \bigcup_{j \in T_i} A_j$. In contrast, define $V[T_i]$ to be the set of vertices appearing in $T_i$, not necessarily with their parents.

Whenever $|S_{i,j}|$ is small and $|T_1| \approx |T_2|$, then $S_{i,j}$ can be efficiently used in algorithm 3. We now devise a dynamic programming algorithm for computing the optimal graph using the cluster tree separators. First, let us root the cluster tree at an arbitrary $U_0 \in U$, inducing an order on the tree vertices. Each cluster $U_j \in U$ is the root of a subtree $T_j$, spanning away from $U_0$. $S_j$ is the tree separator, separating $T_j$ from the rest of $T$ ( $S_0 = \emptyset$). The *sub-vertices* of $U_j$ are its neighbors in $T_j$.

Define for each cluster $U_j$ and each total order $\sigma$ on $S_j$ the weight $W[U_j, \sigma]$ of the maximal partial solution which respects $\sigma$

$$W[U_j, \sigma] = \max_{\substack{\text{acyclic } G \subset H[T_S] \\ \text{respecting } \sigma}} W_{A[T_j]}[G]. \quad (1)$$

The crux of the algorithm is that finding these weights can be done in a recursive manner, based on previously computed maxima.

**Proposition 5.7:** *For each cluster $U_j \in U$ and order $\sigma$ over $S_j$: Let $U_1, \ldots, U_k$ be the sub-vertices of $U_j$. Then $W[U_j, \sigma]$ is equal to*

$$\max_{\sigma'} ( \max_{\substack{\text{acyclic } G \subset H[A_j] \\ \text{respecting } \sigma'}} W_{A_j}[G] + \sum_{i=1}^{k} W[U_i, \sigma'|_{S_i}] )$$

*where $\sigma'$ ranges on all orders on $U_j$ that are consistent with $\sigma$, and $\sigma'|_{S_i}$ is the restriction of $\sigma'$ to an order over $S_i$.*

Proposition 5.7 facilites rapid evaluation of all the tables $W[U, \sigma]$ in one phase, working our way from the leaves inwards towards $U_0$. At the end of this traversal, we have computed the weight of each ordering on all separators adjacent to the root cluster $U_0$. A second phase then traverses $T$ from the root outwards, in order to back-trace the choices made during the first phase, leading to the maximum total weight $W_H[G]$.

Examining the complexity of this algorithm, we see that each cluster $U_j$ is visited twice, the first (more expensive) visit requiring $O(|U_j|! \cdot |A_j| \cdot 2^k)$ operations, where $k$ is the size of the candidate sets. Thus, we get the following result:

**Theorem 5.8:** *If $c$ is the size of the largest cluster in the cluster tree, then finding $G$ that maximizes $W[G]$ can be done in $O(2^k \cdot (c + 1)! \cdot |J|)$.*

In summary, the algorithm is *linear* in the size of the cluster tree but worse than exponential in the size of the largest cluster in the tree.

The discussion until now assumed a fixed cluster tree. In practice we also need to select the cluster tree. This is a well-known and hard problem that is beyond the scope of this paper. However, we note that if there is a small cluster tree, then it can be found in polynomial time [2].

| Method | Iter | Time | Score | KL | Stats |
|---|---|---|---|---|---|
| Greedy | | 40 | -15.35 | 0.0499 | 2656 |
| Disc 5 | 1 | 14 | -18.41 | 3.0608 | 908 |
| | 2 | 19 | -16.71 | 1.3634 | 1063 |
| | 3 | 23 | -16.21 | 0.8704 | 1183 |
| Disc 10 | 1 | 20 | -15.53 | 0.2398 | 1235 |
| | 2 | 26 | -15.43 | 0.1481 | 1512 |
| | 3 | 32 | -15.43 | 0.1481 | 1733 |
| Shld 5 | 1 | 14 | -17.50 | 2.1675 | 915 |
| | 2 | 29 | -17.25 | 1.8905 | 1728 |
| | 3 | 36 | -16.92 | 1.5632 | 1907 |
| Shld 10 | 1 | 20 | -15.86 | 0.5357 | 1244 |
| | 2 | 35 | -15.50 | 0.1989 | 1968 |
| | 3 | 41 | -15.50 | 0.1974 | 2109 |
| Score 5 | 1 | 12 | -15.94 | 0.6756 | 893 |
| | 2 | 27 | -15.34 | 0.0550 | 1838 |
| | 3 | 34 | -15.33 | 0.0479 | 2206 |
| Score 10 | 1 | 17 | -15.54 | 0.2559 | 1169 |
| | 2 | 30 | -15.31 | 0.0352 | 1917 |
| | 3 | 34 | -15.31 | 0.0352 | 2058 |

Table 1: Summary of results on synthetic data from alarm domain. These results report the quality of the network, measured both in terms of the score (BDe score divided by number of instances), and KL divergence to the generating distribution. The other columns measure performance both in terms of execution time (seconds) and the number of statistics collected from the data. The methods reported are **Disc** – discrepancy measure, **Shld** – shielding measure, and **Score** – score based measure.

### 5.6 Cluster-Tree Heuristics

Although the algorithm of the previous section is linear in the number of clusters, it is worse than exponential in the size of the largest cluster. Thus, in many situations we expect it to be hopelessly intractable. Nonetheless, this algorithm provides some intuition on how to decompose the heuristic search for our problem.

The key idea is that although after computing a cluster tree, many of the clusters might be large, we can use a mixture of the exact algorithm on small clusters and heuristic searches such as greedy hill climbing on the larger clusters. Due to space constraints, we only briefly outline the main ideas of this approach.

When $U_j$ is sufficiently small, we can efficiently store the tables $W[U_j, \sigma]$ used by the exact cluster tree algorithm. However, if the clusters are large, then we cannot do the maximization of Proposition 5.7. Instead, we perform a heuristic search, such as greedy hill-climbing, over the space of parents for vertices in $A_j$ to find a partial network that is consistent with the ordering induced by the current assignment.

By proceeding in this manner, we approximate the exact algorithm. This approximation examines a series of small search spaces, that are presumably easier to deal with than the original search space. This approach can be easily extended to deal with cluster trees in which only some of the separators are small.

## 6 Experimental Evaluation

In this section we illustrate the effectiveness of the sparse candidate algorithm. We examine both a synthetic example and a real-life dataset. Our current experiments are designed to evaluate the effectiveness of the general scheme and to show the utility of various measures for selecting candidates in the **Restrict** phase. In the experiments described here we use greedy hill-climbing for the **Maximize** phase. We are currently working on implementation of the heuristic algorithms described in Section 5, and we hope to report results. Some statistics about strongly connected component sizes are reported.

The basic heuristic search procedure we use is a greedy hill-climbing that considers local moves in the form of edge addition, edge deletion, and edge reversal. At each iteration, the procedure examines the change in the score for each possible move, and applies the one that leads to the biggest improvement. These iterations are repeated until convergence. In order to escape local maxima, the procedure is augmented with a simple version of TABU search. It keeps a list of the $N$ last candidates seen, and instead of applying the best local change, it applies the best local change that results in a structure not on the list. Note that because of the TABU list, the best allowed change might actually reduce the score of the current candidate. We terminate the procedure after some fixed number of changes failed to result in an improvement over the best score seen so far. After termination, the procedure returns the best scoring structure it encountered.

In the reported experiments we use this greedy hill-climbing procedure both for the Maximize phase of the sparse candidate algorithm, and as a search procedure by itself. In the former case, the only local changes that are considered are those allowed by the current choice of candidates. In the latter case, the procedure considers all possible local changes. This latter case serves as a reference point against which we compare our results. In the expanded version of this paper, we will also compare to other search procedures.

To compare these search procedures we need to measure both their performance in the task at hand, and their computational cost.

The evaluation of quality is based on the score assigned to the network found by each algorithm. In addition, for synthetic data, we can also measure the true error with respect to the generating distribution. This allows us to assess the significance of the differences between the scores during the search.

Evaluating the computational cost is more complicated. The simplest approach is to measure running time. We report running times on an unloaded Pentium II 300mhz machines running Linux. These running times, however, depend on various coding issues in our implementation. We attempted to avoid introducing bias within our code for either procedure, by using the same basic library for evaluating the score of candidates and for computing and caching of sufficient statistics. Moreover, the actual search is carried by the same code for greedy-hill climbing procedure.

As additional indication of computational cost, we also measured the number of sufficient statistics computed from the data. In massive datasets these computations can be the most significant portion of the running time. To minimize the number of passes over the data we use a cache that allows us to use previously computed statistics, and to

214    Friedman, Nachman, and Peér

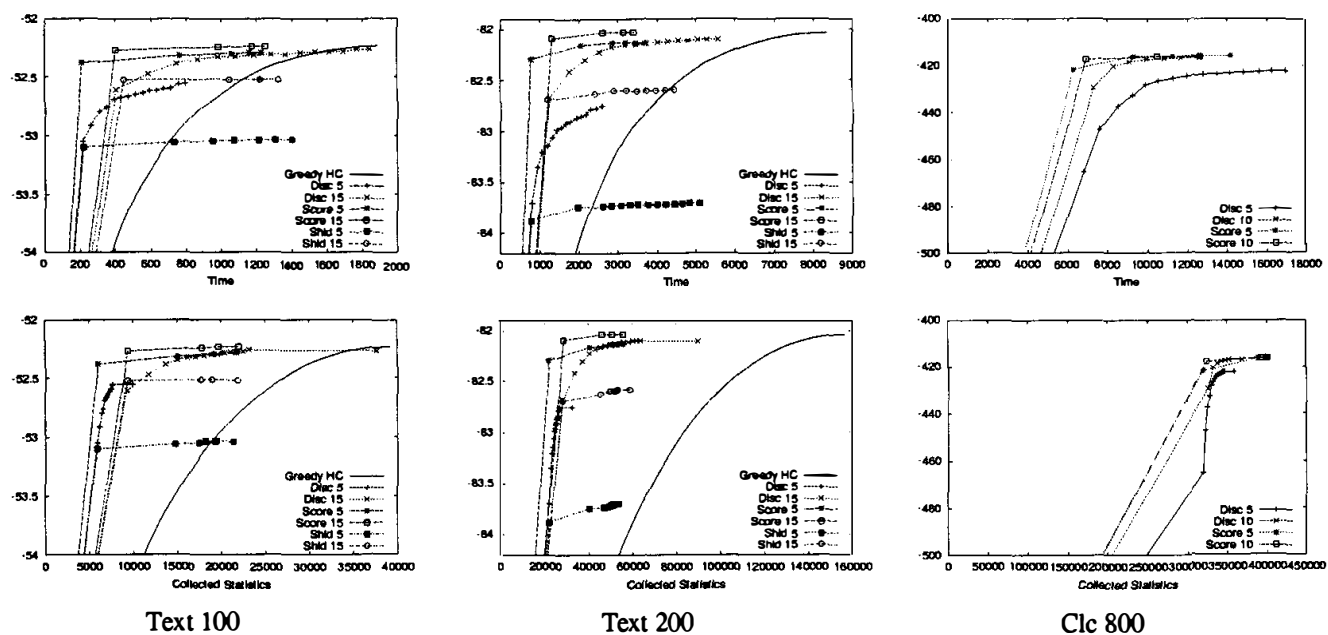

Text 100       Text 200       Clc 800

Figure 4: Graphs showing the performance of the different algorithms on the text and biological domains. The graphs on the top row show plots of score ($y$-axis) vs. running time ($x$-axis). The graphs on the bottom row show the same run measured in terms of score ($y$-axis) vs. number of statistics computed ($x$-axis). The reported methods vary in terms of the candidate selection measure (**Disc** – discrepancy measure, **Shld** – shielding measure, **Score** – score based measure) and the size of the candidate set (k = 10 or 15). The points on each curve for the sparse candidate algorithm are the end result of an iteration.

marginalize statistics to get the statistics of subsets. We report the number of actual statistics that were computed from the data.

Finally, in all of our experiments we used the BDe score of [16] with a uniform prior with equivalent sample size of ten. This choice is a fairly unformed prior that does not code initial bias toward the correct network. The strength of the equivalent sample size was set prior to the experiments and was not tuned.

In the first set of experiments we used a sample of 10000 instances from the "alarm" network [1]. This network has been used for studies of structure learning in various papers, and is treated as a common benchmark in the field. This network contains 37 variables, of which 13 have 2 values, 22 have 3 values, and 2 have 4 values. We note that although we do not consider this data set particularly massive, it does allow us to estimate the behavior of our search procedure. In the future we plan to use synthetic data from larger networks.

The results for this small data set are reported in Table 1. In this table we measure both the score of the networks found and their error with respect to generating distributions. The results on this toy domain show that our algorithm, in particular with the Score selection heuristic, finds networks with comparable score to the one found by greedy hill climbing. Although the timing results for this small scale experiments are not too significant, we do see that the sparse candidate algorithm usually requires fewer statistics records. Finally, we note that the first iteration of the al-gorithm finds reasonably high scoring networks. Nonetheless, subsequent iterations improve the score. Thus, the re-estimation of candidate sets based on our score does lead to important improvements.

To test our learning algorithms on more challenging domains we examined data from text. We used the data set that contains messages from 20 newsgroups (approximately 1000 from each) [18]. We represent each message as a vector containing one attribute for the newsgroup and attributes for each word in the vocabulary. We constructed data sets with different numbers of attributes by focusing on subsets of the vocabulary. We did this by removing common stop words, and then sorting words based on their frequency in the whole data set. The data sets included the group designator and the 99 (*text 100* set) or 199 (*text 200* set) most common words. We trained on 10,000 messages that were randomly selected from the total data set.

The results of these experiments are reported in figure 4. As we can see, in the case of 100 attributes, by using the Score selection method with candidate sets of sizes 10 or 15, we can learn networks that are reasonably close to the one found by greedy hill-climbing in about half the running time and half the number of sufficient statistics. When we have 200 attributes, the speedup is larger than 3. We expect that as we consider data sets with larger number of attributes, this speedup ratio will grow.

To test that, we devised another synthetic dataset, which originates in real biological data. We used gene expression data from [23]. The data describes expression level of 800 cell-cycle regulated genes, over 76 experiments. We learned a network from this dataset, and then sampled 5000 instances from the learned network. We then used this synthetic dataset. See [13] for further details.

The results are reported in figure 4. In these experiments, the greedy hill-climbing search stopped because of lack of memory to store the collected statistics. At that stage it was far from the range of scores shown in the figure. If we try to assess the time it would take it to reach the score of the networks found by the other methods, it seems at least 3 times slower, even by conservative extrapolation. We also note that the discrepancy measure has a slower learning curve than the score measure. Note that after the first iteration, where the initial $O(n^2)$ statistics are collected, each iteration adds only a modest number of new statistics, since we only calculate the measure for pairs of variables that initially had a significant mutual information.

## 7 Conclusion

The contributions of this paper are two fold.

First, we propose a simple heuristic for improving search efficiency. By restricting our search to examine only a small number of candidate parents for each variable, we can find high-scoring networks efficiently. Furthermore, we showed that we can improve the choice of the candidates by taking into account the network we learned, thus getting higher scoring networks. We demonstrated both of these effects in our experiments. These results show that our procedure can lead to dramatic reduction in the learning time. This comes with small loss of quality, at worse, and sometimes can lead to higher scoring networks.

Second, we showed that by restricting each variable to a small group of candidate parents, we can sometimes get theoretical guarantees on the complexity of the learning algorithm. This result is of theoretical interest: to the best of our knowledge, this is the first non-trivial case for which one can find a polynomial time learning algorithm for networks with in-degree greater than one. This theoretical argument might also have practical ramifications. As we showed, even if the exact polynomial algorithm is too expensive, we can use it as a guide for finding good approximate solutions. We are in the process of implementing this new heuristic strategy and evaluating it.

In addition to the experimental results we describe here, our algorithm is already applied in other ongoing works. In [4], the sparse candidate method is combined with the structural EM procedure for learning structure from incomplete data. In that setup, the cost of finding statistics is much higher, since instead of counting number of instances, we have to perform inference for each of the instances. As a consequence the reduction in the number of requested statistics (as shown in our results) leads to significant saving in run time. Similar cost issues occur in [14], where a variant of our algorithm is used for learning probabilistic models from relational databases. Finally, this procedure is a crucial component in our ongoing work in analysis of real-life *gene expression data* that contains thousands of attributes [13].

There are several directions for future research. Our ultimate aim is to use this type of algorithm for learning in domains with thousands of attributes. In such domains the cost of the **Restrict** step of our algorithm is prohibitive (since it is quadratic in the number of variables). We are currently examining heuristic methods for finding good candidates. Once we learn a network based on these candidates, we can use it to help focus on other variables that should be examined in the next **Restrict** step. Another direction of interest is the combination of our methods with other recent ideas for efficient learning from large datasets, such as [20].

## Acknowledgments

This work was supported through the generosity of the Michael Sacher Trust.

## References

[1] I. Beinlich, G. Suermondt, R. Chavez, and G. Cooper. The ALARM monitoring system. In *Proc. 2'nd Euro. Conf. on AI and Med.*. 1989.

[2] H. Bodlaender. A linear-time algorithm for finding tree-decompositions of small treewidth. *SIAM J. Computing*, 25(6):1305–1317, 1996.

[3] R. Bouckaert. *Bayesian Belief Networks: From Construction to Inference*. PhD thesis, Utrecht University, The Netherlands, 1995.

[4] X. Boyen, N. Friedman, and D. Koller. Discovering the structure of complex dynamic systems. In *UAI '99*. 1999.

[5] D. M. Chickering. A transformational characterization of equivalent Bayesian network structures. In *UAI '95*, pp. 87–98. 1995.

[6] D. M. Chickering. Learning Bayesian networks is NP-complete. In *AI&STAT V*, 1996.

[7] D. M. Chickering. Learning equivalence classes of Bayesian network structures. In *UAI '96*, pp. 150–157. 1996.

[8] C. K. Chow and C. N. Liu. Approximating discrete probability distributions with dependence trees. *IEEE Trans. on Info. Theory*, 14:462–467, 1968.

[9] G. F. Cooper and E. Herskovits. A Bayesian method for the induction of probabilistic networks from data. *Machine Learning*, 9:309–347, 1992.

[10] T. H. Cormen, C. E. Leiserson, and R. L. Rivest. *Introduction to Algorithms*. 1990.

[11] T. M. Cover and J. A. Thomas. *Elements of Information Theory*. 1991.

[12] K. J. Ezawa and T. Schuermann. Fraud/uncollectable debt detection using a Bayesian network based learning system. In *UAI '95*, pp. 157–166. 1995.

[13] N. Friedman, I. Nachman, and D. Peér. On using bayesian networks to analyze whole-genome expression data. TR CS99-6 Hebrew University, 1999. In progress. See www.cs.huji.ac.il/labs/pmai2.

[14] L. Getoor, N. Friedman, D. Koller, and A. Pfeffer. Learning probabilistic relational models. In *IJCAI '99*. 1999.

[15] D. Heckerman. A tutorial on learning with Bayesian networks. In M. I. Jordan, editor, *Learning in Graphical Models*, 1998.

[16] D. Heckerman, D. Geiger, and D. M. Chickering. Learning Bayesian networks: The combination of knowledge and statistical data. *Machine Learning*, 20:197–243, 1995.

[17] F. V. Jensen. *An introduction to Bayesian Networks*, 1996.

[18] T. Jochims. A probabilistic analysis of the rocchio algorithm with tfidf for text categorization. In *ICML*, 1997.

[19] W. Lam and F. Bacchus. Learning Bayesian belief networks: An approach based on the MDL principle. *Comp. Int.*, 10:269–293, 1994.

[20] A. W. Moore and M. S. Lee. Cached sufficient statistics for efficient machine learning with large datasets. *J. AI. Res.*, 8:67–91, 1997.

[21] J. Pearl and T. S. Verma. A theory of inferred causation. In *KR '91*, pp. 441–452. 1991.

[22] M. Sahami. Learning limited dependence bayesian classifiers. pp. 335–338, 1996.

[23] P. T. Spellman, G. Sherlock, M. Q. Zhang, V. R. Iyer, K. Anders, M. B. Eisen, P. O. Brown, D. Botstein, and B. Futcher. Comprehensive identification of cell cycle-regulated genes of the yeast sacccharomyces cerevisiae by microarray hybridization. *Mol. Biol. Cell*, 9:3273–3297, 1998.

[24] P. Spirtes, C. Glymour, and R. Scheines. *Causation, prediction, and search*, 1993.